\documentclass[10pt,twocolumn,letterpaper]{article}

\usepackage[pagenumbers]{cvpr} % To force page numbers, e.g. for an arXiv version

\newcommand{\red}[1]{{\color{red}#1}}

\usepackage{nicefrac}  
\definecolor{cvprblue}{rgb}{0.21,0.49,0.74}
\usepackage[pagebackref,breaklinks,colorlinks,allcolors=cvprblue]{hyperref}
\usepackage{graphicx}
\usepackage[accsupp]{axessibility}

\usepackage{amsmath}
\usepackage{amssymb}
\usepackage{booktabs}
\usepackage{times}
\usepackage{epsfig}
\usepackage{amssymb}
\usepackage{enumitem}  % Add this line in the preamble of your document

\usepackage{multirow}
\usepackage{multicol}
\usepackage{siunitx}
\usepackage{tabularx}
\usepackage{array}
\usepackage{comment}
\usepackage{bm}
\usepackage{tikz}
\usepackage{makecell}
\usepackage{pifont}% http://ctan.org/pkg/pifont
\usepackage{fancyhdr}
\usepackage{arydshln}
\usepackage{verbatim}
\usepackage{pifont}
\usepackage{footmisc}
\usepackage{float}
\usepackage{algorithm, algorithmic}
\definecolor{lightblue}{rgb}{0.6, 0.6, 0.8}

\usepackage{microtype}
\def\paperID{2795} % *** Enter the Paper ID here
\def\confName{CVPR}
\def\confYear{2025}

\title{Filter Images First, Generate Instructions Later: 

Pre-Instruction Data Selection for Visual Instruction Tuning}

\author{Bardia Safaei\textsuperscript{1*}, Faizan Siddiqui\textsuperscript{2}, Jiacong Xu\textsuperscript{1}, Vishal M. Patel\textsuperscript{1}, Shao-Yuan Lo\textsuperscript{2}
\and 
\textsuperscript{1}Johns Hopkins University, \textsuperscript{2}Honda Research Institute USA\\
{\tt\small \{bsafaei1, jxu155, vpatel36\}@jhu.edu} \hspace{5pt} {\tt\small \{faizan\_siddiqui, shao-yuan\_lo\}@honda-ri.com}
}

\hypersetup{
    colorlinks=true,        % Enable colored links
    linkcolor=red,          % Set color for internal links (sections, figures, etc.)
}

\begin{document}
\maketitle

\begin{abstract}
Visual instruction tuning (VIT) for large vision-language models (LVLMs) requires training on expansive datasets of image-instruction pairs, which can be costly. Recent efforts in VIT data selection aim to select a small subset of high-quality image-instruction pairs, reducing VIT runtime while maintaining performance comparable to full-scale training. However, a major challenge often overlooked is that generating instructions from unlabeled images \footnote{Throughout the paper, we refer to images without corresponding textual instructions as unlabeled images.} for VIT is highly expensive. Most existing VIT datasets rely heavily on human annotations or paid services like the GPT API, which limits users with constrained resources from creating VIT datasets for custom applications. To address this, we introduce \underline{Pre}-Instruction Data \underline{Sel}ection (\texttt{PreSel}), a more practical data selection paradigm that directly selects the most beneficial unlabeled images and generates instructions only for the selected images. \texttt{PreSel} first estimates the relative importance of each vision task within VIT datasets to derive task-wise sampling budgets. It then clusters image features within each task, selecting the most representative images with the budget. This approach reduces computational overhead for both instruction generation during VIT data formation and LVLM fine-tuning. By generating instructions for only 15\% of the images, \texttt{PreSel} achieves performance comparable to full-data VIT on the LLaVA-1.5 and Vision-Flan datasets. The link to our project page:
\url{https://bardisafa.github.io/PreSel}
\def\thefootnote{*}\footnotetext{This work was mostly done when B. Safaei was an intern at HRI-USA.}
\end{abstract}    
\section{Introduction} \label{sec:intro}

\begin{figure}[t!]
    \begin{center}
        \includegraphics[width=0.96\columnwidth, , trim={0.5cm 0.5cm 0.5cm 0.5cm}]{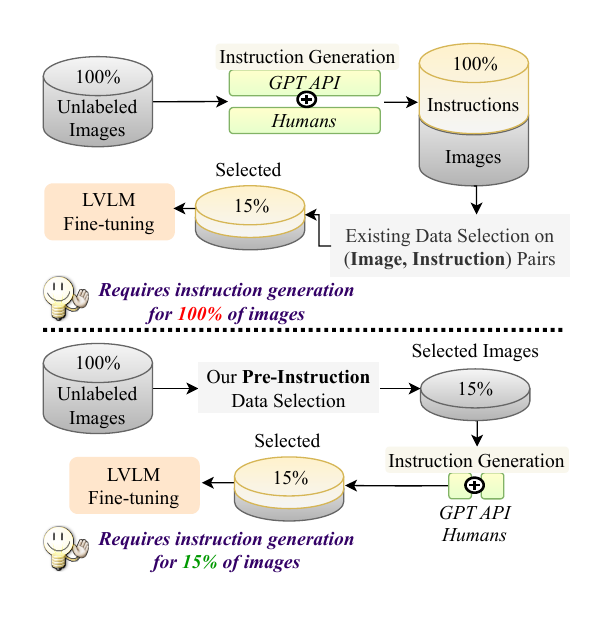}
    \end{center}
    \vskip -15.0pt
    % SY's edit
    \caption{\textbf{Top:} Existing VIT data selection methods assume access to well-prepared VIT datasets in which all the images are already annotated with instructions by costly resources, such as GPT API and human labor. These methods require information on both images and their instructions. \textbf{Bottom:} Our approach performs selection directly on unlabeled images and then utilizes resources to generate instructions exclusively for the selected images. Hence, we not only enable faster fine-tuning but also significantly reduce instruction generation costs (e.g., 15\%).}    
    \label{fig:motive} 
    \vskip -15.0pt
\end{figure}

Recent advances in large vision-language models (LVLMs)~\cite{zhu2023minigpt, bai2023qwen, liu2024improved, dai2023instructblip, liu2024visual} have demonstrated remarkable capabilities in complex multimodal reasoning and human instruction following. Visual instruction tuning (VIT) is a crucial stage in training LVLMs, enabling these models to follow instructions and generalize across various tasks. VIT datasets~\cite{xu2024vision, liu2024improved, yin2024lamm} contain visual instructions from a diverse range of vision tasks, such as visual question answering (VQA)~\cite{goyal2017making, schwenk2022okvqa, marino2019ok}, optical character recognition (OCR)~\cite{mishra2019ocr}, visual grounding~\cite{kazemzadeh2014referitgame}, captioning~\cite{sidorov2020textcaps}, etc. A typical VIT sample is an image-instruction pair consisting of an unlabeled image and instructions generated by human labor or paid services such as the GPT API~\cite{achiam2023gpt, brown2020language,peng2023instruction}. However, two key challenges arise with the current VIT process: (1) Combining visual instructions from diverse tasks could lead to data redundancy, largely increasing training time while yielding only marginal performance gains. (2) Generating high-quality instructions is highly expensive. Specifically, synthesizing instructions from unlabeled images via services like GPT-4~\cite{achiam2023gpt} incurs significant API costs. Furthermore, human involvement is often required to ensure instruction quality, such as in image-grounded question generation, response consistency, and removal of factually incorrect questions.

Recently, several data selection methods for VIT~\cite{lee2024concept, chen-etal-2024-vision, liu2024less, wei2023instructiongpt} have attempted to solve the data redundancy issue by selecting a small subset of informative image-instruction pairs and fine-tuning LVLMs on this subset only, thereby reducing training time. Still, the high-cost issue of instruction generation for large-scale VIT datasets remains overlooked. These methods assume access to well-prepared VIT datasets in which all the images are already annotated with instructions, and their data selection algorithms require information on both images and the corresponding instructions. Hence, these methods are less useful when users need to create new VIT datasets for their custom applications but have limited resources to generate high-quality instructions. To address this, we pose the following question: \textit{``Given a large-scale collection of unlabeled images from multiple vision tasks, how can we select the most impactful images for LVLM fine-tuning before the costly instruction generation step?''} Answering this question would allow us to allocate resources to generate instructions only for the selected images.

In this paper, we propose a novel \underline{Pre}-instruction data \underline{Sel}ection approach, termed \texttt{PreSel}, a more practical data selection paradigm that not only reduces dataset size for faster fine-tuning but also significantly cuts down instruction generation costs (see Figure \ref{fig:motive}). \texttt{PreSel} consists of two stages: (1) task-importance estimation and (2) task-wise cluster-based selection. In a typical VIT dataset, the quantity of data is unbalanced across different vision tasks, and this quantity distribution can greatly impact LVLM performance~\cite{dai2024cotbal, lee2024concept}. Unlike prior works that use heuristic methods to balance task data \cite{tong2024cambrian, radford2021learning, xu2023demystifying}, our Stage~1 introduces an automated task-importance estimation process. It begins by randomly sampling a small fraction of unlabeled images from each task (e.g., as little as $5\%$) and obtaining instructions for them. Using this small reference set, we propose the \textit{Instruction Relevance Score (IRS)} to compute the relative importance of each task. The importance values determine task-wise sampling budgets for data selection. Next, Stage~2 employs a lightweight vision encoder (e.g., DINOv2~\cite{oquab2023dinov2}) to extract features for the unlabeled images in each task, and performs clustering to select the most representative images within each cluster. Finally, instructions are acquired for the selected images, preparing them for LVLM fine-tuning.

We verify the effectiveness of \texttt{PreSel} on two large-scale VIT datasets: LLaVA-1.5 \cite{liu2024improved} and Vision-Flan \cite{xu2024vision}. Our experiments assume that these datasets consist solely of unlabeled images to simulate the more practical pre-instruction selection scenario. Using only 15\% of the VIT data, \texttt{PreSel} achieves performance comparable to LVLMs fine-tuned on the full dataset. Notably, this efficiency is attained by pre-instruction data selection, which requires only 15\% of the instruction generation cost. We also demonstrate the transferability of \texttt{PreSel}’s selection to LVLMs across various architectures and sizes. We release our selected subset to the public to facilitate efficient model development. To the best of our knowledge, we are the first to introduce the pre-instruction VIT data selection paradigm, operating on unlabeled images before instruction generation. The key contributions are summarized as follows.

\begin{itemize}
    \vspace{-0mm}
    \setlength{\itemsep}{0pt}
    \setlength{\topsep}{0pt}
    \setlength{\parsep}{0pt}
    \setlength{\partopsep}{0pt}
    \item We are the first to introduce the pre-instruction VIT data selection paradigm, a more practical data selection paradigm that reduces not only VIT runtime but also instruction generation cost.
    
    \item We propose \texttt{PreSel}, a novel pre-instruction data selection approach that operates on unlabeled images before instruction generation.
    
    \item Experiments on LLaVA-1.5 and Vision-Flan show that, using only 15\% of the data and instruction generation cost, \texttt{PreSel} achieves performance comparable to LVLMs fine-tuned on the full VIT dataset.
\end{itemize}

\section{Related Work}
\label{sec:related}

\medskip\noindent\paragraph{Visual Instruction Tuning.}
\vspace{-8mm}
Instruction tuning is an essential training step that enables large language models (LLMs) to follow instructions and generalize across various tasks~\cite{chiang2023vicuna,dubey2024llama,achiam2023gpt,peng2023instruction}. Recent advances, such as LLaVA~\cite{liu2024visual,liu2024improved}, MiniGPT-4~\cite{zhu2023minigpt}, InstructBLIP~\cite{dai2023instructblip} and Qwen-VL-Chat~\cite{bai2023qwen}, extend this technology from the natural language processing (NLP) domain to multi-modality, i.e., VIT for fine-tuning LVLMs. These works include large-scale VIT data, which demands considerable LVLM training time. Moreover, to create such large-scale VIT datasets, these works mostly employ GPT-family models~\cite{achiam2023gpt, brown2020language,peng2023instruction} to synthesize visual instructions from unlabeled images, with human involvement often needed to ensure instruction quality. This instruction generation process is highly expensive.

\medskip\noindent\paragraph{Data Selection for Visual Instruction Tuning.}
\vspace{-5mm}
Due to high training costs, recent studies have investigated data-efficient instruction tuning~\cite{li2023quantity, xia2024less, zhou2023lima}. LIMA~\cite{zhou2023lima} is the first to show that training LLMs on a selected small subset of data can reach same-level performance. LESS~\cite{xia2024less} designs an optimizer-aware algorithm to estimate data influences. Li et al.~\cite{li2023quantity} develop the Instruction-Following Difficulty (IFD) metric to select essential data.

Several recent attempts further explore data selection for VIT \cite{lee2024concept, liu2024less, chen-etal-2024-vision, wei2023instructiongpt}. InstructionGPT-4~\cite{wei2023instructiongpt} uses metrics such as CLIP- and GPT-based scores to select high-quality VIT samples, but its fine-tuning results are suboptimal. Self-Filter~\cite{chen-etal-2024-vision} first trains a score-net along with LVLM fine-tuning on all VIT data and then uses the score-net to select a data subset for a second round of VIT. However, this double-round process increases the overall training cost, contradicting the motivation of data selection. Liu et al.~\cite{liu2024less} leverage gradients to estimate the importance of each sample, yet the gradient information is expensive as it requires back-propagating the target LVLM. COINCIDE~\cite{lee2024concept} introduces a clustering method to identify concept-skill compositions for LVLMs, but it needs instructions as references in its process.

While these VIT data selection methods achieve comparable performance using only a small portion of data, their selection pipelines require full access to image-instruction pairs. In contrast, our approach performs selection directly on unlabeled images, generating instructions only for the selected ones. Hence, we not only enable faster fine-tuning but also significantly reduce instruction generation costs.

\section{Methodology}
In this section, we first formulate the problem of pre-instruction data selection for visual instruction tuning and then elaborate on the proposed approach.

% \begin{figure*}
%     \centering
%     \includegraphics[width=0.65\textwidth]{sec/pics/cvpr_bar_graph_with_highlights.pdf}  % Use \textwidth to span both columns
%     \caption{An illustration of $Q$, $R$, and $I$ in a VIT sample.}
%     \label{fig:method_qir} 
% \end{figure*}

\medskip\noindent
\textbf{Problem Formulation.}
Consider a large pool of unlabeled images \(\mathcal{D}\) assembled from various datasets to construct a VIT dataset with \(M\) distinct vision tasks \(\{T_i\}_{i=1}^M\), where \(\mathcal{D} = \bigcup_{i=1}^{M} T_i\), and we denote the number of samples in \(\mathcal{D}\) as \(|\mathcal{D}|\). Examples of vision tasks include VQA, OCR, etc., and each task \(T_i\) consists of a set of unlabeled images \(T_i = \{I_a^{i}\}_{a=1}^{|T_i|}\). Note that tasks may overlap in images, i.e., \(T_i \cap T_j \neq \emptyset\) for some \(i \neq j\). For an unlabeled image \(I\) from task \(T_i\), the corresponding textual instruction \(Y\) is generated as \(Y = F_i(I)\), where \(F_i\) represents the instruction generation process for \(T_i\). Note that \(F_i\) is not a straightforward mathematical function; rather, it is a costly, task-specific procedure, potentially involving resources such as the GPT API~\cite{achiam2023gpt, brown2020language} or human annotators who label images with instructions based on defined guidelines.

The goal of pre-instruction data selection is to select a small subset of highly beneficial unlabeled images $\mathcal{D}_S \subset \mathcal{D}$, where \(|\mathcal{D}_S| \ll \left| \mathcal{D} \right|\), and then it only acquires instructions for this small subset. Fine-tuning an LVLM on the resulting image-instruction pairs, \(\left\{ (I_a, Y_a) \right\}_{a=1}^{\left| \mathcal{D}_S\right|}\), should maximally improve the LVLM's instruction-following capabilities and achieve performance comparable to full-scale fine-tuning on \(\mathcal{D}\) with complete instructions: \(\left\{ (I_a, Y_a) \right\}_{a=1}^{\left| \mathcal{D}\right|}\). The key difference between the pre-instruction data selection paradigm and existing VIT data selection methods is that the latter assumes access to instructions of all images (i.e., \(\left\{ (I_a, Y_a) \right\}_{a=1}^{\left| \mathcal{D}\right|}\)), while pre-instruction data selection solely relies on unlabeled images \(\left\{ I_a \right\}_{a=1}^{\left| \mathcal{D}\right|}\) for selecting $\mathcal{D}_S$. Hence, this paradigm enables efficiency in both training and instruction generation.

\begin{figure}[t!]
    \begin{center}
        \includegraphics[width=1\columnwidth, trim={0.3cm 0.3cm 0.3cm 0.2cm}]{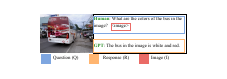}
    \end{center}
    \vspace{-2mm}
    \caption{An illustration of $Q$, $R$, and $I$ in a VIT sample. Instruction $Y = \{Q, R\}$.}    
    \label{fig:method_qir}
    \vspace{-4mm}
\end{figure}

\medskip\noindent
\textbf{Overview.} Our approach begins with a Task-Importance Estimation mechanism to obtain the optimal proportion of each task $T_i$ in $\mathcal{D}_S$. To achieve this, we first randomly select a small reference set of images $\mathcal{D}_{\text{ref}} \subset \mathcal{D}$, where \(\left| \mathcal{D}_{\text{ref}} \right| \ll \left| \mathcal{D}_S \right| \ll \left| \mathcal{D} \right| \), and acquire their corresponding instructions, \(\left\{ (I_a, Y_a) \right\}_{a=1}^{\left| \mathcal{D}_{\text{ref}}\right|}\). Each instruction \(Y_a\) is then decomposed into questions \(Q_a\) and responses \(R_a\) (i.e., $Y_a = \{Q_a, R_a\}$; see Figure \ref{fig:method_qir}), which are used to compute our proposed \textit{Instruction Relevance Score (IRS)} for the samples in \(\mathcal{D}_{\text{ref}}\). The average IRS over images in each task $T_i$ determines the relative proportions of these tasks in $\mathcal{D}_S$, termed $w(T_i)$. Next, we employ a lightweight vision encoder, e.g., DINOv2~\cite{oquab2023dinov2}, to extract visual features for the remaining unlabeled images and cluster them in each task. Finally, given the derived task proportion $w(T_i)$, the most representative images from each cluster are selected via the \textit{Neighbor Centrality (NC)} score.

\begin{figure*}
    \centering
    \includegraphics[width=1\textwidth, trim={1.4cm 0.5cm 0.8cm 0.3cm}]{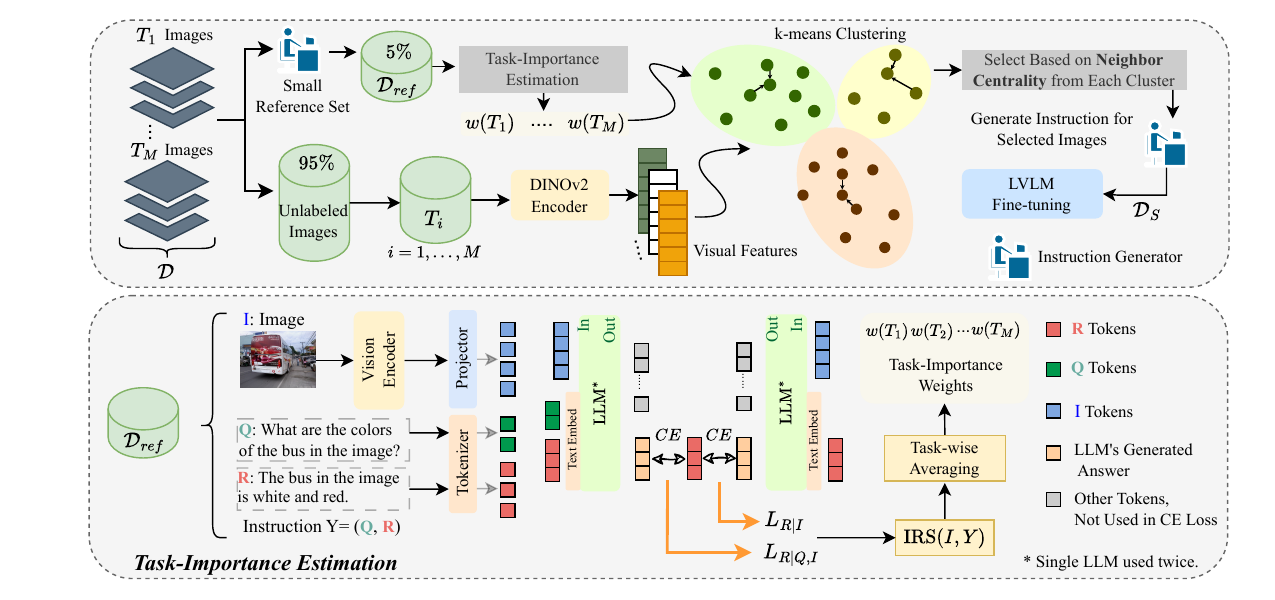}  % Use \textwidth to span both columns
    \caption{We propose $\texttt{PreSel}$, an efficient \underline{Pre}-Instruction Data \underline{Sel}ection approach for Visual Instruction Tuning (VIT). Given a large pool of unlabeled images $D$ from various tasks, $\texttt{PreSel}$ first estimates the importance of each task $T_i$ via a randomly selected small reference set $\mathcal{D}_{ref}$ with instructions generated. Each instruction ($Y$) is split into questions ($Q$) and responses ($R$) to compute the Instruction Relevance Score (IRS), which determines task proportions $w(T_i)$ in the final selected subset $\mathcal{D}_S$. Given the derived task proportions, it then uses the DINOv2 vision encoder to extract features from the remaining unlabeled images, perform clustering within each task, and select representative images using the Neighbor Centrality (NC) score. The collection of selected images from all tasks is assembled as $\mathcal{D}_S$.}
    \label{fig:framework} 
\end{figure*}

% \vspace{-1mm}
\subsection{Task-Importance Estimation}
\label{subsec:task-importance}
Determining an appropriate proportion of samples from each task for the final selected subset $\mathcal{D}_S$ is crucial. Simply relying on the number of images available per task to set these proportions can lead to suboptimal performance, as tasks often differ in their levels of redundancy. Also, some tasks may be effectively learned through training on related tasks, making direct sampling from them less essential \cite{dai2024cotbal, lee2024concept}.

We start by fine-tuning the target LVLM on the image-instruction pairs in a randomly selected small reference set \(\mathcal{D}_{\text{ref}}\), which comprises only a small fraction (e.g., as little as 5\%) of the entire VIT dataset \(\mathcal{D}\). This initial fine-tuning, conducted for one epoch, equips the LVLM with basic instruction-following abilities. We refer to this fine-tuned model as the \textit{reference model}. Inspired by the loss-based data selection methods in NLP \cite{li2023quantity, marion2023less}, we extend the loss-based idea to address the more complicated multimodal scenario and leverage the loss predictions of the reference model on $\mathcal{D}_{ref}$ to define our \textit{Instruction Relevance Score (IRS)} for estimating task importance. 

As shown in Figure \ref{fig:method_qir}, each VIT example in $\mathcal{D}_{ref}$ is represented as a triplet $(I, Q, R)$, where $I$ represents the image, $Q$ the textual question (from a human), and $R$ the response (from GPT). $Q$ and $R$ may extend over multiple interaction rounds. The proposed IRS is calculated by comparing the reference model's next-token cross-entropy (CE) loss with and without the \( Q \) tokens as part of the input. This score evaluates how much the provided \( Q \) contributes to generating the ground-truth response \( R \). 
Formally, the next-token cross-entropy (CE) loss for $R$ given the tokens of $I$ and $Q$ as context is as follows:
\vspace{-3mm}
\begin{equation}
\label{eq:l_rqi}
\mathcal{L}_{R|Q,I} = - \frac{1}{|\mathbf{t}^R|} \sum_{j=1}^{|\mathbf{t}^R|} \log P_\theta\left(t_j^R \mid I, Q, \mathbf{t}_{<j}^R \right), 
\vspace{-1mm}
\end{equation}
where $\mathbf{t}^R$ is the tokenized $R$ with $|\mathbf{t}^R|$ tokens, and $\mathbf{t}_{<j}^R$ is the sequence of tokens preceding the $j$-th token in $R$. $P_\theta$ denotes the predicted probability distribution of the reference model, parameterized by $\theta$. We then calculate the loss without the $Q$ given as context:
\vspace{-3mm}
\begin{equation}
\label{eq:l_ri}
\mathcal{L}_{R|I} = - \frac{1}{|\mathbf{t}^R|} \sum_{j=1}^{|\mathbf{t}^R|} \log P_\theta\left(t_j^R \mid I, \mathbf{t}_{<j}^R \right),    
\end{equation}
where the response is only conditioned on the image context. The proposed IRS is formulated as the ratio of these two losses as follows:
\vspace{-3mm}
\begin{equation}
\label{eq:irs}
    \text{IRS} = \frac{\mathcal{L}_{R|Q,I}}{\mathcal{L}_{R|I}}.
\end{equation}

A higher IRS indicates that adding the \( Q \) context to \( I \) does not assist in refining the model for easier generation of \( R \). In contrast, a lower IRS shows that the model's confusion regarding $R$ is reduced when $Q$ is provided as input, emphasizing the necessity of \( Q \) for VIT. We then compute the average IRS over all samples in $\mathcal{D}_{ref}$ that belong to task $T_i$ as follows:
\vspace{-3mm}
\begin{equation}
\label{eq:s_t}
s(T_i) = \frac{1}{|D_{ref}^i|} \sum_{\substack{I \in T_i}} \text{IRS}(I,Y),
\end{equation}

where $|D_{ref}^i|$ denotes the number of samples in $\mathcal{D}_{ref}$ that belong to $T_i$. Based on the definition of IRS, a lower $s(T_i)$ indicates a higher importance of $T_i$. The final relative proportion of each task within $\mathcal{D}_S$ is defined as:
\vspace{-1mm}
\begin{equation}
\label{eq:w_t}
    w(T_i) = \frac{\exp\left(-s(T_i) / \tau\right)}{\sum_{j=1}^M \exp\left(-s(T_j) / \tau\right)},
\end{equation}
where we set the temperature value $\tau=\frac{1}{\sqrt{M}}$.

\subsection{Task-wise Cluster-based Selection}
After determining the relative proportion of each task using the reference set, we focus on selecting informative unlabeled images within each task for instruction generation. For the unlabeled images in task $T_i$, we first extract their visual features using the pre-trained DINOv2 \cite{oquab2023dinov2} model, a lightweight vision encoder. Specifically, given an input image \( I \in T_i \), we obtain the feature vector \( \mathbf{v}_I \) from the last transformer layer's [$\texttt{CLS}$] token after its Layer Normalization.
\begin{comment}
%($\mathsf{LN}$) as:
% \vspace{-1mm}
\begin{equation}
\label{eq:dino_feats}
\mathbf{v}_I = \mathsf{LN}\left( \mathbf{h}_L^{[\texttt{CLS}]} \right) \in \mathbb{R}^{D},
\vspace{-1mm}
\end{equation}
where \( \mathbf{h}_L^{[\texttt{CLS}]} \) denotes the [\texttt{CLS}] token output from the last transformer layer \( L \), and $D$ is the feature dimension.
\end{comment}
We then cluster these obtained \( \mathbf{v}_I \) features of task $T_i$ into $C$ clusters $\{A_c^i\}_{c=1}^C$ using the k-means \cite{macqueen1967some} algorithm, where we set $C=\frac{|T_i|}{100}$. To select samples from the \( c \)-th cluster \( A_c^i \) within \( T_i \), we consider both its relative size \( |A_c^i| \) and the importance weight \( w(T_i) \) of task \( T_i \). Specifically, we choose:

\begin{equation}
\label{eq:budget}
n_c = \left\lfloor \frac{w(T_i) \cdot |A_c^i|}{|T_i|} \cdot |\mathcal{D}_S| \right\rfloor
\end{equation}
images from cluster \( A_c^i \). This approach ensures a diverse selection of images within each cluster, taking into account its size and the overall importance of the corresponding task.

\medskip\noindent
\textbf{Intra-Cluster Selection.} Within each cluster, we select the \( n_c \) most representative images based on the \textit{Neighbor Centrality (NC)} score, defined as:
\vspace{-1mm}
\begin{equation}
s_{nc}(I) = \frac{1}{k}\cdot\sum_{I_a\in k\text{NN}(I)}\operatorname{sim}(\mathbf{v}_I, \mathbf{v}_{I_a}).
\label{eq:representativeness}
\vspace{-1mm}
\end{equation}
Here we denote the $k$-nearest neighbors of a given image $I$ in feature space as $k$NN$(I)$, and $\operatorname{sim}(\cdot, \cdot)$ is the cosine similarity. 
A higher $s_{nc}$ indicates that the image is closely situated to its neighbors, implying it is more likely to be a representative sample rather than an outlier. 

Finally, the collection of selected images from all tasks is assembled as $\mathcal{D}_S$. We utilize resources to generate instructions only for images in $\mathcal{D}_S$, which are then used to fine-tune the LVLM. Figure \ref{fig:framework} provides an overview.

% Table 1
\begin{table*}
    \renewcommand{\arraystretch}{1.1}
    \centering
    \resizebox{1\textwidth}{!}{$
        \begin{tabular}{l|rr|ccccccccc|c}
        \toprule
        \multirow{2}{*}{Method}
        & Req. & Sel. & VQAv2 & SQA-I & TextVQA & MME & \multicolumn{2}{c}{MMBench} & SEED-Bench & MM-Vet & POPE & \textbf{Rel.}\\
        \cmidrule(lr){8-9}
        & Inst.& Inst.& & & & & en & cn & & & & (\%) \\
        \hline
        BLIP-2~{\scriptsize\textcolor{darkgray}{[ICML-23]}} \cite{li2023blip} & 100\% & - & 65.0 & 61.0 & 42.5 & 1293.8 & - & - & 49.7 & 22.4 & 85.3 & - \\
        InstructBLIP-7B~{\scriptsize\textcolor{darkgray}{[NeurIPS-23]}} \cite{dai2023instructblip} & 100\% & 1.2M & - & 60.5 & 50.1 & - & 36.0 & 23.7 & 58.8 & 26.2 & - & - \\
        Shikra~{\scriptsize\textcolor{darkgray}{[Arxiv-23]}} \cite{chen2023shikra} & 100\% & 5.5M & 77.4 & - & - & - & 58.8 & - & - & - & - & - \\
        IDEFICS-80B~{\scriptsize\textcolor{darkgray}{[HuggingFace-23]}} \cite{idefics2023} & 100\% & 1M & 60.0 & - & 30.9 & - & 54.5 & 38.1 & 53.2 & - & - & - \\
        Qwen-VL-Chat~{\scriptsize\textcolor{darkgray}{[Arxiv-23]}} \cite{bai2023qwen} & 100\% & 50M & 78.2 & 68.2 & 61.5 & 1487.5 & 60.6 & 56.7 & 65.4 & - & - & - \\
        InstructionGPT-4~{\scriptsize\textcolor{darkgray}{[Arxiv-23]}} \cite{wei2023instructiongpt} & 100\% & 0.2K & - & - & 20.6 & 463.3 & 31.4 & - & - & - & - & - \\
        \hline
        \hline
        \rowcolor{brown!10} LLaVA-1.5-7B & 100\% & 665K & 79.1 & 68.4 & 57.9 & 1417.6 & 66.0 & 58.9 & 66.8 & 30.0 & 87.5 & 100 \\
        \hline
        Random & \textbf{15\%} & 93K & 75.3 & 67.8 & 54.3 & 1397.5 & 61.0 & 53.5 & 62.4 & \underline{30.2} & 84.9 & 95.7 \\
        TypiClust~{\scriptsize\textcolor{darkgray}{[ICML-22]}} \cite{hacohen2022active} & \textbf{15\%} & 93K & \underline{76.0} & \underline{68.2} & 53.3 & 1396.2 & \underline{64.3} & \textbf{57.1} & \underline{62.8} & 29.7 & 85.6 & \underline{96.8} \\
        CLIP-Score~{\scriptsize\textcolor{darkgray}{[ICML-21]}} \cite{radford2021learning} & 100\% & 93K & 71.7 & 64.5 & 53.4 & 1380.3 & 51.0 & 48.0 & 56.2 & 29.9 & 84.0 & 90.3 \\
        EL2N~{\scriptsize\textcolor{darkgray}{[NeurIPS-21]}} \cite{paul2021deep} & 100\% & 93K & \textbf{76.1} & 66.5 & 50.2 & 1405.2 & 58.2 & 48.5 & 61.8 & 30.0 & 83.3 & 93.0 \\
        Perplexity~{\scriptsize\textcolor{darkgray}{[Arxiv-23]}} \cite{marion2023less} & 100\% &93K & 73.1 & 67.2 & 54.6 & 1283.7 & 54.7 & 47.8 & 58.3 & \textbf{30.7} & 85.5 & 91.9 \\
        IFD ~{\scriptsize\textcolor{darkgray}{[NAACL-24]}} \cite{li2023quantity} & 100\% & 93K & 74.0 & 66.5 & 51.8 & 1307.2 & 57.2 & 50.6 & 59.4 & 28.1 & \underline{86.6} & 91.8 \\
        Self-Filter ~{\scriptsize\textcolor{darkgray}{[ACL-24]}} \cite{chen-etal-2024-vision} & 100\% & 93K & 74.0 & 62.3 & 51.4 & 1356.5 & 48.1 & 45.4 & 56.3 & 29.0 & \textbf{87.0} & 88.8 \\
        COINCIDE~{\scriptsize\textcolor{darkgray}{[EMNLP-24]}} \cite{lee2024concept} & 100\% & 93K & \textbf{76.1} & 67.7 & \underline{54.8} & \underline{1414.9} & 60.5 & 53.9 & 62.0 & 28.5 & 86.4 & 95.5 \\
        \hline
         \rowcolor{purple!6} \textbf{\texttt{PreSel}} (Ours) & \textbf{15\%} & 93K & 75.0 & \textbf{70.1} & \textbf{55.2} & \textbf{1457.7} & \textbf{64.8} & \underline{56.5} & \textbf{63.8} & 29.6 & 85.4 & \textbf{97.9}\\
        \bottomrule
        \end{tabular}
    $}
    \vspace{-2mm}
    \caption{\textbf{Results on the \textit{LLaVA-1.5} dataset}. We compare \texttt{PreSel} with several data selection approaches across multiple multimodal benchmarks. ``Sel. Inst.'' indicates the number of visual instructions used to fine-tune the LVLM. Our experiments involve selecting 15\% of the full VIT dataset (93K samples) for fine-tuning. ``Req. Inst.'' shows percentage of images for which instructions are generated. \texttt{PreSel} only requires instruction generation for the selected images (15\%), whereas other methods need instructions for 100\% of the data to perform selection. The best result is \textbf{bolded} and the runner-up is \underline{underlined}.}
    \label{tab:llava}
    \vspace{-4mm}
\end{table*}

\section{Experiments}
\subsection{Experimental Setup}
\medskip\noindent
\textbf{Datasets.}
We conduct experiments on two large-scale VIT datasets: LLaVA-1.5 \cite{liu2024improved} and Vision-Flan \cite{xu2024vision}. LLaVA-1.5, a widely used dataset, comprises more than 600K samples across about 10 different tasks. Vision-Flan includes 191 distinct tasks and 186K data points. Evaluating on this dataset allows us to assess performance when dealing with a high number of tasks and relatively fewer images per task. It is important to note that although our approach can be used to efficiently construct new VIT datasets or scale up existing ones, we use these established VIT datasets for evaluation purposes. Specifically, we perform data selection on the unlabeled images and treat the corresponding instructions for the selected images as the generated instructions.

\medskip\noindent
\textbf{Baselines.}
We compare our proposed approach with several data selection methods, namely Random, CLIP-Score \cite{radford2021learning}, TypiClust \cite{hacohen2022active}, EL2N \cite{paul2021deep}, Perplexity \cite{marion2023less}, IFD \cite{li2023quantity}, Self-Filter \cite{chen-etal-2024-vision}, and COINCIDE \cite{lee2024concept}. In particular, Self-Filter and COINCIDE are recent state-of-the-art VIT data selection methods that are closest to our setting. IFD, EL2N, and Perplexity are state-of-the-art selection methods used for instruction tuning in the NLP domain. Due to the similarity of our paradigm to active learning (AL) \cite{li2024survey,Safaei_2025_WACV,xie2023active, safaei2024entropic}, we also transplant and adapt TypiClust, a strong clustering-based AL method, as it cannot be directly applied in our setting. See Section~\red{A4} in Appendix for more details of these methods.

\medskip\noindent
\textbf{Evaluation Benchmarks.}
We evaluate the performance of different fine-tuned LVLMs across several multimodal evaluation benchmarks, including VQAv2 \cite{goyal2017making}, ScienceQA \cite{lu2022learn}, TextVQA \cite{singh2019towards}, MME-Perception \cite{yin2023survey}, MMBench \cite{liu2025mmbench}, MMBench (Chinese version) \cite{liu2025mmbench}, SEED-Bench \cite{li2023seed}, MM-Vet \cite{yu2023mm}, and POPE \cite{li2023evaluating}. This diverse set of metrics effectively assesses the quality of LVLMs fine-tuned with samples selected by the compared methods, covering aspects such as object hallucination, open-ended short answers, scientific question answering, text-intensive VQA, robust yes/no and multiple-choice QA, and visual conversations. 

\medskip\noindent
\textbf{Implementation details.}
Unless stated otherwise, we set the sampling ratio to 15\% of the VIT dataset in all our comparisons and fine-tune the LLaVA-1.5-7B~\cite{liu2024improved} pre-training checkpoint \footnote{This checkpoint is taken after pre-training the projector for feature alignment but before the VIT stage.: \url{https://huggingface.co/liuhaotian/llava-v1.5-mlp2x-336px-pretrain-vicuna-7b-v1.5}.} (the model before LLaVA's VIT stage) with the selected VIT samples. The proposed method requires a small, randomly selected reference set $\mathcal{D}_{ref}$ for task-importance estimation, which is equal to 5\% of the VIT dataset. At a 15\% sampling ratio, we select an additional 10\% of data using our method and combine it with the 5\% reference set $\mathcal{D}_{ref}$ to ensure fair comparisons. As the scale of different evaluation benchmarks varies, we report the \textit{Average Relative Performance (Rel. \%)} in our comparisons. The relative performance for each benchmark is calculated as $$\frac{\text{The model's performance}}{\text{The full fine-tuned model's performance}} \times \text{100}.$$
In all experiments, we fine-tune the LVLM for one epoch using LoRA \cite{hu2021lora}. We use the same training details as the official LLaVA-1.5 paper \cite{liu2024improved}. We conduct all experiments using four H100 (80 GB) GPUs.
\subsection{Main Results}
\noindent
\textbf{High performance with minimal instruction generation.} Table \ref{tab:llava} presents the results on the LLaVA-1.5 dataset, comparing $\texttt{PreSel}$ with various data selection baselines to highlight its efficiency. We set the sampling ratio to 15\% of the full LLaVA-1.5 size. It can be observed that the proposed method significantly outperforms other baselines in average relative performance and achieves comparable performance to full-scale LLaVA fine-tuning. Notably, this performance gain is achieved with instruction generation for only the selected 15\% of images, as shown in the \textit{Required Instructions} column. Interestingly, random selection outperforms many baselines, even though these methods have access to all instructions prior to selection. This may be because these methods often select a poor distribution of samples across tasks, negatively impacting performance. Also, these methods tend to overly prioritize sample difficulty for generation, neglecting the visual diversity of the selected samples. Our explicit task-importance estimation coupled with representative sample selection from diverse clusters yields a clear performance boost over other baselines. In particular, $\texttt{PreSel}$ outperforms Random by 2.2\% and the second-best performing baseline by 1.1\%, showing its effectiveness.

\medskip\noindent
\textbf{$\texttt{PreSel}$ is robust across varying task diversities.} As the LLaVA-1.5 dataset contains a relatively limited number of tasks (approximately 10), we also conduct experiments on Vision-Flan, which covers a diverse set of 191 vision tasks, as shown in Table \ref{tab:vision-flan}. The results demonstrate that $\texttt{PreSel}$ even slightly surpasses the performance of the full-scale fine-tuned LLaVA model with only 15\% generated instructions. This verifies that the proposed approach can robustly adapt to VIT datasets with varying numbers of vision tasks without performance degradation. Particularly, our results on Vision-Flan highlight significant efficiency advantages for creating VIT datasets with a diverse set of tasks, because (1) the task-importance estimation step can automatically identify the important tasks to focus on, and (2) our approach performs selection directly on unlabeled images, generating instructions only for the selected ones. This reduced instruction generation cost enables a more efficient expansion of VIT datasets.

\begin{figure}
    \centering
    \includegraphics[width=1\linewidth]{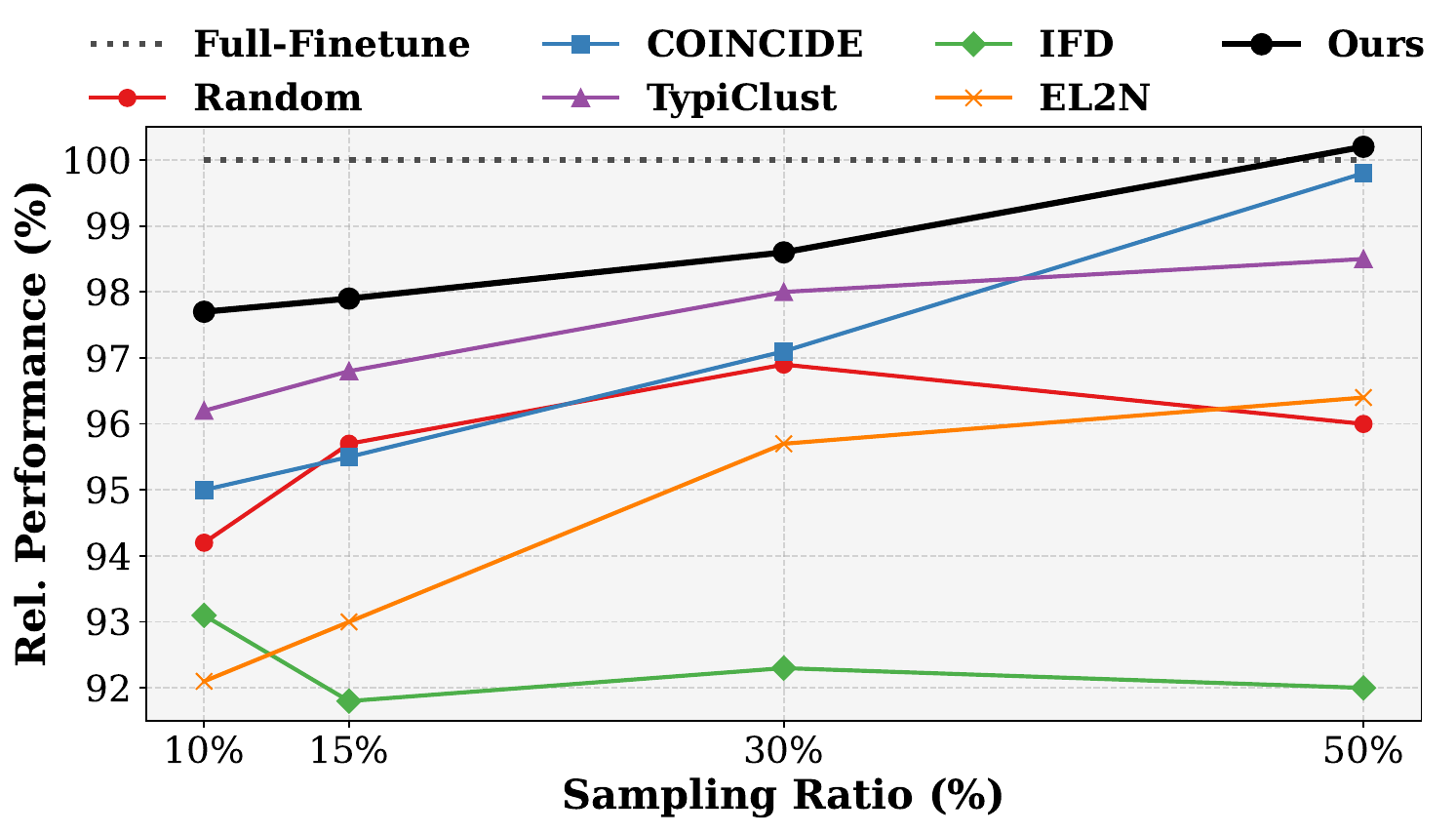}
    \vspace{-3mm}
    \caption{Average relative performance of data selection methods on LLaVA-1.5 at different sampling ratios.}
    \label{fig:curve}
    \vspace{-4mm}
\end{figure}

% Table 2
\begin{table*}
    \renewcommand{\arraystretch}{1.1}
    \centering
    \resizebox{1\textwidth}{!}{$
        \begin{tabular}{l|rr|ccccccccc|c}
        \toprule
        \multirow{2}{*}{Method}
        & Req. & Sel. & VQAv2 & SQA-I & TextVQA & MME & \multicolumn{2}{c}{MMBench} & SEED-Bench & MM-Vet & POPE & \textbf{Rel.}\\
        \cmidrule(lr){8-9}
        & Inst.& Inst.& & & & & en & cn & & & & (\%) \\
        \hline
        \rowcolor{brown!10} LLaVA-1.5-7B & 100\% & 186K & 69.6 & 55.6 & 38.3 & 1238.1 & 53.4 & 48.2 & 55.6 & 27.7 & 85.7 & 100 \\
        \hline
        Random & \textbf{15\%} & 28K & \textbf{66.5} & 62.1 & 38.7 & 1238.6 & 43.6 & 43.1 & 48.3 & 28.1 & 83.0 & 96.1 \\
        TypiClust~{\scriptsize\textcolor{darkgray}{[ICML-22]}} \cite{hacohen2022active} & \textbf{15\%} & 28K & 65.8 & 59.2 & 37.7 & 1194.1 & 32.4 & 45.1 & 44.9 & 28.2 & 81.6 & 92.0 \\
        CLIP-Score~{\scriptsize\textcolor{darkgray}{[ICML-21]}} \cite{radford2021learning} & 100\% & 28K & 63.0 & 62.3 & 39.7 & 1058.0 & 37.5 & 44.2 & 44.3 & 28.0 & 82.0 & 92.2 \\
        EL2N~{\scriptsize\textcolor{darkgray}{[NeurIPS-21]}} \cite{paul2021deep} & 100\% & 28K & 63.6 & 62.7 & \underline{42.4} & \underline{1253.8} & 44.7 & 37.7 & 49.2 & 27.1 & 79.5 & 95.2 \\
        Perplexity~{\scriptsize\textcolor{darkgray}{[Arxiv-23]}} \cite{marion2023less} & 100\% &28K & \underline{66.1} & 53.3 & 39.5 & 1195.0 & 25.7 & 39.8 & 38.8 & \textbf{29.3} & 83.5 & 88.2 \\
        IFD ~{\scriptsize\textcolor{darkgray}{[NAACL-24]}} \cite{li2023quantity} & 100\% & 28K & 65.0 & 57.8 & 42.0 & 1210.9 & 30.4 & 40.8 & 39.1 & 26.9 & 82.6 & 90.0 \\
        Self-Filter ~{\scriptsize\textcolor{darkgray}{[ACL-24]}} \cite{chen-etal-2024-vision} & 100\% & 28K & 64.9 & 59.3 & \textbf{42.6} & \textbf{1262.2} & 42.1 & 43.8 & 44.5 & 25.1 & 80.9 & 94.2 \\
        COINCIDE~{\scriptsize\textcolor{darkgray}{[EMNLP-24]}} \cite{lee2024concept} & 100\% & 28K & 66.0 & \underline{63.9} & 33.0 & 1184.4 & \underline{49.6} & \textbf{48.2} &\textbf{53.9} & 26.1 & \textbf{84.3} & \underline{97.1} \\
        \hline
         \rowcolor{purple!6} \textbf{\texttt{PreSel}} (Ours) & \textbf{15\%} & 28K & 64.1 & \textbf{66.2} & 39.7 & 1218.8 & \textbf{50.4} &\underline{45.4} & \underline{53.5} & \underline{29.1} & \underline{84.1} &\textbf{100.1}\\
        \bottomrule
        \end{tabular}
    $}
    \vspace{-2mm}
    \caption{\textbf{Results on the \textit{Vision-Flan} dataset}. We compare \texttt{PreSel} with several data selection approaches across multiple multimodal benchmarks. ``Sel. Inst.'' indicates the number of visual instructions used to fine-tune the LVLM. Our experiments involve selecting 15\% of the full VIT dataset (28K samples) for fine-tuning. ``Req. Inst.'' shows the percentage of images for which instructions are generated. \texttt{PreSel} only requires instruction generation for the selected images (15\%), whereas other methods need instructions for 100\% of the data to perform the selection. The best result is \textbf{bolded}, and the runner-up is \underline{underlined}.}
    \label{tab:vision-flan}
    \vspace{-1mm}
\end{table*}

% Table 3
\begin{table*}[t]
    \centering
    \tiny
    \renewcommand{\arraystretch}{1}
    \resizebox{1\textwidth}{!}{$
    \begin{tabular}{lccccc}
    \toprule
    Methods & Selection Cost & Finetuning Cost & Inst. Gen. Cost & Total Cost& \textbf{Rel. (\%)} \\
    \midrule
    \rowcolor{brown!10} Full Finetune & -- &76.0 GPU-hr & 100\%$\cdot C$ & 76.0 GPU-hr + 100\%$\cdot C$ & 100 \\
    Self-Filter~\cite{chen-etal-2024-vision} & 73.5 GPU-hr & 11.0 GPU-hr & 100\%$\cdot C$ & 84.5 GPU-hr + 100\%$\cdot C$& 88.8 \\
    COINCIDE~\cite{lee2024concept} & 55.5 GPU-hr & 11.0 GPU-hr& 100\%$\cdot C$ & 66.5 GPU-hr + 100\%$\cdot C$& 95.5 \\ \rowcolor{purple!6}
    \textbf{\texttt{PreSel} (Ours)} & 9.0 GPU-hr & 11.0 GPU-hr & 15\%$\cdot C$& \textbf{20.0 GPU-hr + 15\%$\cdot C$} & \textbf{97.9} \\
    \bottomrule
    \end{tabular}
    $}
    \caption{Comparison of VIT costs for \texttt{PreSel}, other VIT data selection methods, and full-scale LVLM fine-tuning. GPU-hr refers to one H100 (80 GB) GPU used for one hour, and $C$ is the total cost for instruction generation across all unlabeled VIT images.}
    \label{tab:wall_clock}
    \vspace{-4mm}
\end{table*}
\subsection{Detailed Analyses}
\label{sec:detailed analysis}
\begin{figure}[t!]
% \vspace{-6mm}
    \begin{center}
        \includegraphics[width=1\columnwidth, trim={0.3cm 1.7cm 0.2cm 0.1cm}]{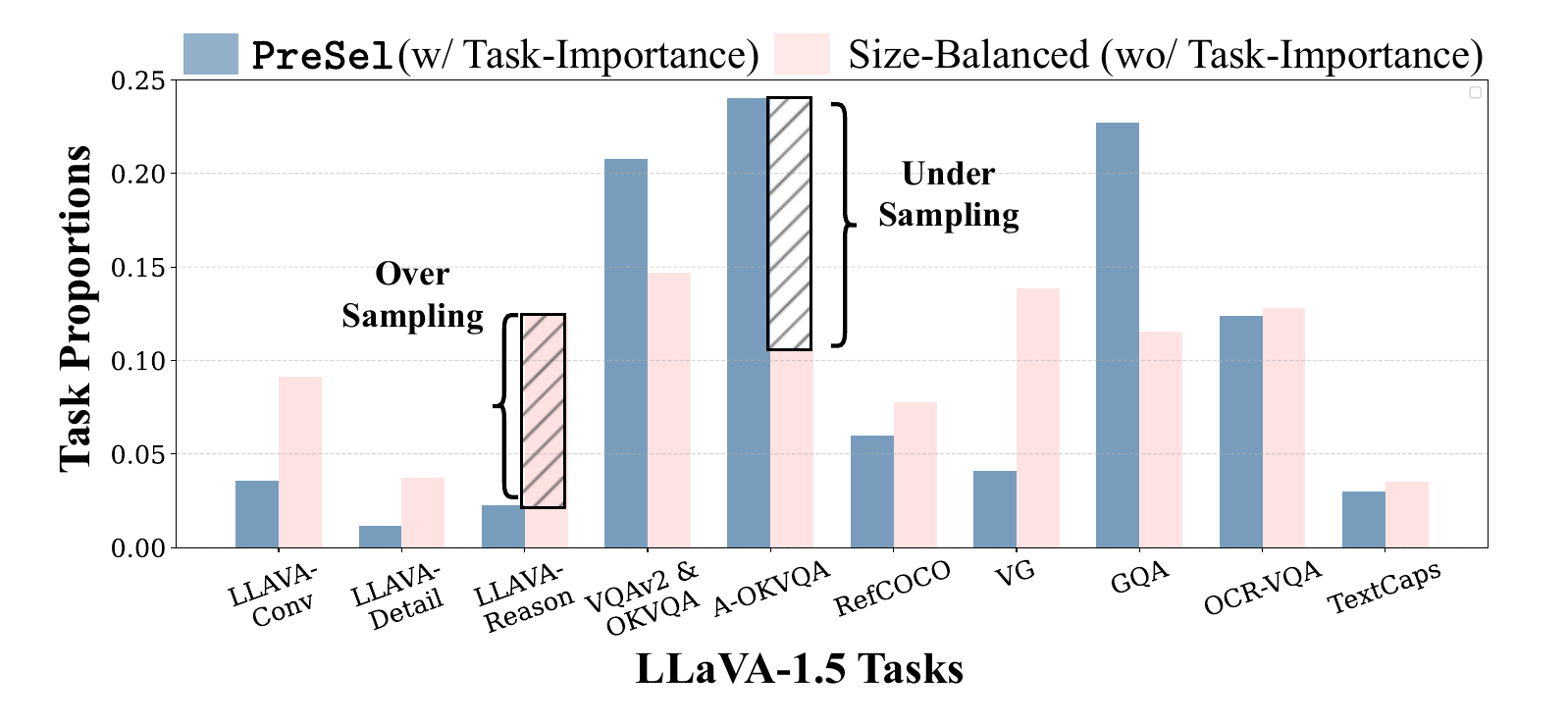}
    \end{center}
    \vspace{-2mm}
    \caption{A demonstration of task proportions for the LLaVA-1.5 dataset assigned by \texttt{PreSel} and Size-Balanced sampling.}    
    \label{fig:task_prop} 
    \vspace{-2mm}
\end{figure}

\textbf{Performance across various sampling ratios.} Figure \ref{fig:curve} shows the average relative performance of different approaches on LLaVA-1.5 for sampling ratios from 10\% to 50\% of the full dataset. 
Interestingly, the performance of random selection begins to degrade beyond a certain ratio, underscoring that merely adding more data does not always lead to improved VIT performance. Hence, effective data selection is crucial for identifying and incorporating only the most beneficial samples. It can be seen that most data selection methods show performance gains with larger sampling budgets; however, IFD fails to improve. We discovered this occurs because IFD predominantly selects samples from the LLaVA-v1 \cite{liu2024visual} subset of the LLaVA-1.5 dataset (tasks: LLaVA-Conv,-Detail,-Reason), which comprise almost 25\% of the entire dataset. Both our experiments and prior research \cite{xu2024vision} indicate that only a few samples from these tasks are needed for strong performance. Consequently, IFD inefficiently allocates its budget to LLaVA-v1, missing more informative samples from other critical tasks. Our task-importance estimation approach effectively tackles this challenge (see Figure \ref{fig:task_prop}). Overall, $\texttt{PreSel}$ surpasses the full-finetuning baseline at 50\% sampling ratio and consistently delivers superior performance across all ratios.
\begin{table*}[t]
    \centering
    % First Table (Balancing Ablation moved here)
    \begin{minipage}[t]{0.32\linewidth}
        \centering
        \renewcommand{\arraystretch}{0.8}
        \begin{tabular}{clc}
        \toprule
        \textbf{NC}& \textbf{Balancing } & \textbf{Rel. (\%)} \\
        \midrule
        \multirow{3}{*}{\checkmark}& Uniform & 95.8 \\
        & Size-Balanced & 97.1 \\
        & Task-Importance & \textbf{97.9} \\
        \midrule
        \multirow{3}{*}{--}&Uniform & 94.8 \\
        &Size-Balanced & 94.9 \\
        &Task-Importance & \textbf{96.9} \\
        \bottomrule
        \vspace{-5mm}
        \end{tabular}
        \caption{The effect of different task balancing strategies on the performance for the LLaVA-1.5 dataset. Ours is Task-Importance.}
        \label{tab:balancing_ablation}
    \end{minipage}%
    \hfill
    % Second Table (Key Components moved here)
    \begin{minipage}[t]{0.36\linewidth}
        \centering
        \renewcommand{\arraystretch}{1.2}
        \begin{tabular}{ccc}
        \toprule
        \textbf{NC} & \textbf{Clustering} & \textbf{Rel. (\%)} \\
        \midrule
        -- & -- & 96.2 \\
        -- & \checkmark & 96.9 \\
        \checkmark & -- & 97.2 \\
        \checkmark & \checkmark & \textbf{97.9} \\
        \bottomrule
        \vspace{-7mm}
        \end{tabular}
        \caption{Ablation study on NC and k-means clustering. The experiments are conducted on the LLaVA-1.5 dataset with the sampling ratio set to 15\%.}
        \label{tab:nc_cluster}
    \end{minipage}%
    \hfill
    % Third Table
    \begin{minipage}[t]{0.26\linewidth}
        \centering
        \renewcommand{\arraystretch}{0.9}
        %\resizebox{0.9\linewidth}{!}{%
        \begin{tabular}{cc}
        \toprule
        \textbf{\# of Nearest} & \textbf{Rel.} \\ 
        \textbf{Neighbors ($k$)} & \textbf{(\%)} \\ \midrule
        5 & 97.9 \\
        10 & 98.2 \\
        20 & 97.4 \\
        \midrule
        Second-best & 96.8 \\
        \bottomrule
        \vspace{-5mm}
        \end{tabular}
        %} % End of resizebox
        \caption{Sensitivity to $k$. The experiments are conducted on LLaVA-1.5 with the sampling ratio set to 15\%.}
        \label{tab:hyperparam}
    \end{minipage}
    \vspace{-4mm}
\end{table*}

\medskip\noindent
\textbf{Cost analysis.} Table \ref{tab:wall_clock} compares the computational cost of \texttt{PreSel} with two recent VIT data selection methods and full-scale fine-tuning. Using a 15\% sampling ratio on the LLaVA-1.5 dataset, \texttt{PreSel} achieves Pareto efficiency improvements in data selection. It selects 15\% of images in just 9 GPU hours, whereas COINCIDE and Self-Filter require 55.5 and 73.5 GPU hours, respectively, for 15\% of image-instruction pairs. COINCIDE's high cost stems from TinyLLaVA-2B forward passes on all image-instruction pairs, while \texttt{PreSel} requires DINOv2 forward passes only for images without instructions and LLaVA-1.5-7B forward passes for just 5\% of image-instruction pairs (two per sample for IRS computation). The computational cost of \texttt{PreSel} scales linearly with dataset size, ensuring scalability. Self-Filter is inefficient as it requires training an LVLM on all image-instruction pairs before selection.

Additionally, \texttt{PreSel} significantly reduces instruction generation costs by selecting data directly from unlabeled images, whereas COINCIDE and Self-Filter require generating instructions for all images. In the table, instruction generation costs are denoted as $C$, a highly expensive process. Overall, \texttt{PreSel} achieves substantial efficiency gains in both GPU usage and instruction generation while maintaining competitive performance with full-scale fine-tuning.

\medskip\noindent
\textbf{The effect of our Task-Importance Estimation.} In Table \ref{tab:balancing_ablation}, we present experiments on the LLaVA-1.5 dataset with sampling ratio set to 15\% to evaluate the effectiveness of the proposed \textit{Task-Importance Estimation} component within our framework. Specifically, we conduct two sets of experiments. In the top three rows of the table, we use the intra-cluster selection approach \textit{Neighbor Centrality (NC)}, which is the default in $\texttt{PreSel}$. To completely decouple the effect of our NC component on task-importance estimation, we use Random for intra-cluster selection in the bottom three rows. \textit{Uniform} indicates that we select the same number of images from each task in the VIT dataset, regardless of the size or importance of a particular task; that is, each task will have an equal number of selected samples in $\mathcal{D}_S$. \textit{Size-Balanced} denotes that the budget for each task is set proportional to its relative size compared to other tasks in the VIT dataset, so larger tasks will have more samples selected in $\mathcal{D}_S$. \textit{Task-Importance} refers to our proposed approach, where the budget for each task is allocated based on its estimated importance, as described in Section \ref{subsec:task-importance}. The results in Table \ref{tab:balancing_ablation} confirm the effectiveness of our proposed Task-Importance Estimation approach. Specifically, sampling based on Task-Importance outperforms Size-Balanced sampling by significant margins of 0.8\% and 2.0\%, with and without the NC component, respectively. Furthermore, we observe that Size-Balanced sampling performs better than Uniform. This is expected, as the size of each task in the LLaVA-1.5 dataset was heuristically adjusted to achieve reasonably good performance during the creation of this VIT dataset \cite{liu2024improved}. Consequently, Size-Balanced sampling is not entirely arbitrary or random. However, in practice, it is highly expensive to heuristically determine such a size balance across a large number of vision tasks when creating a VIT dataset for a custom application. Our task-importance estimation mechanism effectively automates this step. 

In Figure \ref{fig:task_prop}, we compare the task proportions assigned by our method to those assigned by Size-Balanced on the LLaVA-1.5 dataset. We observe that Size-Balanced assigns almost 25\% of the total budget to the LLaVA-v1 \cite{liu2024visual} tasks (LLaVA-Conv, -Detail, -Reason). However, our experiments and previous research \cite{lee2024concept,xu2024vision} have shown that a small fraction of these tasks is sufficient for good VIT performance. In contrast, \texttt{PreSel} automatically reduces the allocation to these tasks and assigns more budget to tasks such as A-OKVQA and GQA, which are more integral for VIT \cite{lee2024concept, dai2024cotbal}.

\medskip\noindent
\textbf{The effect of Neighbor Centrality (NC) and clustering.} In Table \ref{tab:nc_cluster}, we conduct ablation studies on LLaVA-1.5 with the sampling ratio set to 15\% to verify the effectiveness of the proposed k-means clustering within each task and intra-cluster selection based on NC scores. In the first row, we perform random selection on the images within each task based on the budget, without involving any clustering. In the second row, we apply clustering to each task's images and randomly sample from each cluster according to the budget defined in Equation \ref{eq:budget}. The third row is similar to the first, but we select images from each task based on the NC score. Finally, the last row represents the full version of \texttt{PreSel}, which includes both the clustering and NC components. The results confirm the positive contribution of each component to the overall performance, with the complete \texttt{PreSel} achieving the best results.

\medskip\noindent
\textbf{Sensitivity to the number of nearest neighbors ($k$).} In our NC score calculations, we use a hyperparameter $k$, which represents the number of nearest samples in feature space considered for computing the NC score. In Table \ref{tab:hyperparam}, we repeat the experiment on LLaVA-1.5 with the sampling ratio set to 15\% for $k=5,10,20$ and evaluate the performance change. We observe that \texttt{PreSel} is quite robust to variations in the value of $k$, consistently achieving higher average relative performance than the second-best method across all values of $k$. We set $k$ to 5 in all our experiments.

\section{Conclusion}
\label{sec:conclusion}
In this paper, we introduce a new data selection paradigm, termed pre-instruction data selection for VIT, which aims to reduce both VIT runtime and instruction generation costs. We propose \texttt{PreSel}, an effective pre-instruction data selection approach that operates directly on unlabeled images before instruction generation. \texttt{PreSel} leverages a novel Task-Importance Estimation mechanism to automatically identify the most impactful vision tasks for budget allocation. It then selects the most beneficial images from each task for instruction annotation, according to the budgets. Our extensive experiments on the LLaVA-1.5 and Vision-Flan datasets demonstrate that \texttt{PreSel} significantly outperforms other state-of-the-art data selection baselines and achieves performance comparable to full-scale fine-tuning of LVLMs, while requiring instructions for only the selected 15\% unlabeled images. \texttt{PreSel} also enables easier creation of multi-task VIT datasets for custom applications under constrained resources.

{
    \small
    \bibliographystyle{ieeenat_fullname}
    \bibliography{main}
}

\clearpage
\maketitlesupplementary
\setcounter{table}{6}

\setcounter{section}{0}
\renewcommand\thesection{A\arabic{section}}

\section{Limitations and Potential Societal Impacts}
\vspace{-6mm}
\noindent\paragraph{Limitations.}  Currently, our approach utilizes the reference instructions for task-importance estimation. However, given the multimodal nature of LVLMs, these instructions may also be leveraged to improve the selection process on unlabeled images. We leave this limitation as an interesting direction for future work.
\vspace{-3mm}
\noindent\paragraph{Potential Negative Societal Impacts.}
The proposed method could potentially enable malicious actors to fine-tune open-source LVLMs more easily for illegal purposes. The integration of machine learning security mechanisms could be studied to mitigate such risks.

\begin{table*}
    \renewcommand{\arraystretch}{1.1}
    \centering
    \resizebox{1\textwidth}{!}{$
        \begin{tabular}{l|l|ccccccccc|c}
        \toprule
        \multirow{2}{*}{Model} & \multirow{2}{*}{Method} & VQAv2 & SQA-I & TextVQA & MME & \multicolumn{2}{c}{MMBench} & SEED-Bench & MM-Vet & POPE & \textbf{Rel.} \\
        \cmidrule(lr){7-8}
        & & & & & & en & cn & & & & (\%) \\
        \midrule
        \multirow{3}{*}{Vicuna-13B \cite{chiang2023vicuna}}
          & Full Finetune & 80.0 & 72.0 & 58.3 & 1400.4 & 68.2 & 60.7 & 68.0 & 34.3 & 87.1 & 100 \\
          & Random & \textbf{76.2} & 69.9 & 56.8 & \textbf{1452.3} & 62.8 & 56.0 & \textbf{65.0} & 33.0 & 86.6 & 96.6 \\
          &  \textbf{\texttt{PreSel}} (Ours) & 76 & \textbf{70.8} & \textbf{57.0} & 1404.3 & \textbf{66.4} & \textbf{60.9} & \textbf{65.0} & \textbf{34.4} & \textbf{87.2} & \textbf{98.3} \\
        \midrule
        \multirow{3}{*}{Llama-3-8B \cite{dubey2024llama}}
          &  Full Finetune & 81.4 & 78.2 & 63.8 & 1561.2 & 75.2 & 73.1 & 71.8 & 36.9 & 85.6 & 100 \\
          & Random & \textbf{77.5} & 78.3 & \textbf{59.8} & 1455.0 & 72.5 & 69.1 & 68.3 & 35.7 & \textbf{84.9} & 96.0 \\
          &  \textbf{\texttt{PreSel}} (Ours) & 77.0 & \textbf{79.0} & 59.2 & \textbf{1501.4} & \textbf{73.3} & \textbf{70.6} & \textbf{68.5} & \textbf{37.7} & 84.6 & \textbf{97.2} \\
        \bottomrule
        \end{tabular}
    $}
    \vspace{-1mm}
    \caption{\textbf{Results for \textit{LLaVA-Vicuna-13B} and \textit{LLaVA-Llama-8B} models.} We report the performance of $\texttt{PreSel}$ across different model architectures and sizes. The experiments are conducted on the LLaVA-1.5 dataset and the sampling ratio is set to 15\% for both Random and $\texttt{PreSel}$ methods. We directly use the samples acquired for the LLaVA-Vicuna-7B model to fine-tune the additional LVLMs in this experiment. The average relative performance across all evaluation benchmarks is reported.}
    \label{tab:arch_size}
    %\vspace{-2mm}
\end{table*}

\section{Different Model Sizes and Architectures}
\texttt{PreSel} adapts to various model sizes and architectures. In addition to the results presented in Table \red{1} and Table \red{2}, which are based on the LLaVA-7B model using Vicuna-7B \cite{chiang2023vicuna} as the LLM, we conduct two additional sets of experiments on the LLaVA-1.5 dataset, as shown in Table \ref{tab:arch_size}. In these experiments, we change the LLM to Vicuna-13B \cite{chiang2023vicuna} and Llama-8B \cite{dubey2024llama} to evaluate the transferability of the samples selected by $\texttt{PreSel}$ across different model sizes and architectures. It is important to note that we directly use the samples selected by $\texttt{PreSel}$ with LLaVA-7B as the reference model to fine-tune the additional LVLMs, without any further processing. From the table, it is evident that our selected samples are highly beneficial for fine-tuning LVLMs with different architectures or sizes. Specifically, $\texttt{PreSel}$ outperforms Random by large margins of 1.7\% and 1.2\% in average relative performance for the LLaVA-Vicuna-13B and LLaVA-Llama-8B models, respectively.

\section{More Analysis on the Size of Reference Set} 
We leverage a small, randomly selected set of image-instruction pairs as the reference set, $\mathcal{D}{\text{ref}}$, to estimate task-importance values during the Task-Importance Estimation stage. By default, the size of $\mathcal{D}_{\text{ref}}$ is set to 5\% of the total size of the VIT dataset, $\mathcal{D}$. In Table \ref{tab:reference_set}, we reduce the reference set size to just 1\% of the total VIT dataset to evaluate the effect of $|\mathcal{D}_{\text{ref}}|$ on the average relative performance. The rest of the experimental settings are identical to those in Table \red{1} of the main paper: we set the sampling ratio to 15\% and conduct experiments on the LLaVA-1.5 dataset. Our experiments show that the estimated task-importance values remain almost identical for both the 1\% and 5\% cases. Consequently, the final performance is robust to the size of the reference set, as shown in Table \ref{tab:reference_set}.

\begin{table*}
    \renewcommand{\arraystretch}{1.1}
    \centering
    \resizebox{1\textwidth}{!}{$
        \begin{tabular}{c|c|ccccccccc|c}
        \toprule
        $\frac{\textstyle |\mathcal{D}_{\text{ref}}|}{\textstyle |\mathcal{D}|}$ & Method & VQAv2 & SQA-I & TextVQA & MME & MMB-E & MMB-C & SEED-Bench & MM-Vet & POPE & \textbf{Rel.} (\%) \\
        \midrule
        - & Full Finetune & 79.1 & 68.4 & 57.9 & 1417.6 & 66.0 & 58.9 & 66.8 & 30.0 & 87.5 & 100 \\
        5\% & \textbf{\texttt{PreSel}} & 75.0 & 70.1 & 55.2 & 1457.7 & 64.8 & 56.5 & 63.8 & 29.6 & 85.4 & 97.9 \\
        1\% & \textbf{\texttt{PreSel}} & 74.9 & 69.7 & 54.2 & 1436.5 & 63.1 & 57.4 & 63.9 & 30.0 & 86.5 & 97.7 \\
        \bottomrule
        \end{tabular}
        $}
    % \vspace{-1mm}
    \caption{The effect of $|\mathcal{D}_{\text{ref}}|$ on average relative performance. The experiments are conducted on the LLaVA-1.5 dataset and the sampling ratio is set to 15\%. ``Full Finetune'' denotes a LLaVA-1.5-7B model fine-tuned on the entire LLaVA-1.5 image-instruction pairs.}
    \label{tab:reference_set}
    % \vspace{-2mm}
\end{table*}

\begin{algorithm}[tb]
	\caption{Our Proposed \texttt{PreSel} Approach}
	\label{alg:presel}
	\begin{algorithmic}[1]
		\STATE \textbf{Input:}
		\STATE \quad unlabeled images \(\mathcal{D} = \bigcup_{i=1}^{M} T_i\) where $T_i$ is the $i$-th 
    
    \quad task, number of vision tasks $M$, a lightweight vision 
    
    \quad encoder (DINOv2), our LVLM, the small randomly 
    
    \quad selected reference set $\mathcal{D}_{ref}$

		\STATE \textbf{Process:}
            \STATE \quad $\mathcal{D}_{S} \gets \o$ \quad \emph{{\textcolor{lightblue}{\# Initialization for the Selected Subset}}}
            \STATE \quad \emph{{\textcolor{lightblue}{\# Task-Importance Estimation}}}

            \STATE \quad for each $(I, Q, R) \in \mathcal{D}_{ref}$: calculate $\mathcal{L}_{R|Q,I}$ and 
            
            \quad $\mathcal{L}_{R|I}$ via Eq. \red{1} and Eq. \red{2} to obtain IRS via Eq. \red{3} 
		\STATE \quad \textbf{for} $i=1,...,M$ \textbf{do}
		  \STATE \quad \quad calculate $s(T_i)$ and $w(T_i)$ via Eq. \red{4} and Eq. \red{5}
        \STATE \quad \emph{\textcolor{lightblue}{\# Task-wise Cluster-based Selection}}
        \STATE \quad \textbf{for} $i=1,...,M$ \textbf{do} 
        \STATE \quad \quad $\mathcal{D}_{T_i} \gets \o$ \quad \quad \emph{\textcolor{lightblue}{\# Selected Images for Task $T_i$}}
        \STATE \quad \quad for unlabeled images in $T_i$, extract visual features 
        
        \quad \quad using DINOv2 encoder
        \STATE \quad \quad cluster visual features in $T_i$ into $C$ clusters 
        
        \quad \quad $\{A_c^i\}_{c=1}^C$ where $C=\frac{|T_i|}{100}$
        \STATE \quad \quad \emph{{\textcolor{lightblue}{\# Intra-Cluster Selection.}}}
        \STATE \quad \quad \textbf{for} $c=1,...,C$ \textbf{do}
        \STATE \quad \quad \quad for $A_c^i$, set the cluster budget $n_c$ via Eq. \red{6}
        
        \STATE \quad \quad \quad calculate $s_{nc}$ for all images within $A_c^i$ via Eq. \red{7}
        \STATE \quad \quad \quad rank images in $A_c^i$ based on $s_{nc}$ and select the 
        
        \quad \quad \quad top $n_c$ samples with the highest values $\mathcal{D}_{T_i}^c$ 
        \STATE \quad \quad \quad $\mathcal{D}_{T_i} \gets \mathcal{D}_{T_i} \cup \mathcal{D}_{T_i}^c$
        \STATE \quad \quad acquire instructions for images in $\mathcal{D}_{T_i}$
        \STATE \quad $\mathcal{D}_{S} \gets \mathcal{D}_{T_1} \cup \mathcal{D}_{T_2}, ... \cup \mathcal{D}_{T_M} \cup \mathcal{D}_{ref}$ 
        \STATE \quad \emph{{\textcolor{lightblue}{\# Visual Instruction Tuning}}}
        \STATE \quad fine-tune the LVLM 
        \STATE \quad \textbf{Return} the fine-tuned LVLM
            
	\end{algorithmic}
 \label{alg:algorithm}
 
\end{algorithm}

\section{Baselines}
\label{sec:baselines}
In this section, we provide details about the baselines with which we compared our method.

\begin{itemize}
    \item \textbf{CLIP-Score \cite{radford2021learning}.} In CLIP-Score, the cosine similarity between the images and their corresponding textual instructions is used for selection. In our experiments, we select samples with high similarity.
    
    \item \textbf{TypiClust \cite{hacohen2022active}.} TypiClust is an active learning approach originally designed for multi-round data selection for image classification under low-budget regimes. It adaptively clusters the unlabeled data and selects the most typical samples from the clusters that have the highest number of unlabeled samples and the fewest already labeled samples. We adapt a single-round version of this method to our setting.
    
    \item \textbf{EL2N \cite{paul2021deep}.} This method uses the Error L2-Norm (EL2N) to quantify a sample's informativeness. Specifically, it calculates the average L2-Norm distance between generated tokens and ground-truth text tokens to produce this score.

    \item \textbf{Perplexity \cite{marion2023less}.} This method uses the average next-token prediction loss as a measure of the LVLM's uncertainty with respect to a sample. Following prior research \cite{ marion2023less}, we select samples with medium score values instead of top values, as it leads to better results.

    \item \textbf{IFD \cite{li2023quantity}.} This method proposes the Instruction-Following Difficulty (IFD) metric to select samples with corresponding instructions that have a minimal impact on the model's loss.

    \item \textbf{Self-Filter~\cite{chen-etal-2024-vision}.} The Self-Filter method involves initially training a score-net while fine-tuning LVLM on the entire VIT dataset. Afterwards, the score-net is employed to choose a subset of data for a subsequent VIT round. Nonetheless, this two-step procedure escalates the total training expense, opposing the purpose of data selection.

    \item \textbf{COINCIDE \cite{lee2024concept}.} This method first utilizes the TinyLLaVA-2B \cite{zhou2024tinyllava}  model to extract both image and instruction features from multiple layers. It then concatenates all features, performs spherical clustering on them, and selects samples from clusters proportional to their overall transferability.
\end{itemize}

\section{ShareGPT Data in LLaVA-1.5 Dataset}
The LLaVA-1.5 dataset includes a task named ShareGPT, comprised of text-only instructions generated by the ShareGPT model. In our experiments, we have excluded these text-only instructions as our primary focus is to select the most beneficial `images' for visual instruction tuning. One can easily choose a subset of ShareGPT data or use all of it along with our selected VIT samples for fine-tuning LVLMs.

\section{Detailed Algorithm for \texttt{PreSel}}
We elaborate on the details of our \texttt{PreSel} pre-instruction data selection approach in Algorithm \ref{alg:algorithm}.

\end{document}